\documentclass[runningheads]{llncs}
\usepackage[T1]{fontenc}
\usepackage{graphicx}
\usepackage{verbatim}
\usepackage{booktabs}
\usepackage{multirow}
\usepackage{xcolor}
\usepackage{makecell}
\usepackage{pifont}
\usepackage{array}
\usepackage{subcaption}
\usepackage{bbm}
\usepackage{amssymb}
\usepackage{amsmath}
\newcommand{\ourmethod}{{\fontfamily{ppl}\selectfont WSI-Agents}}

\begin{document}
\title{\includegraphics[width=0.5cm]{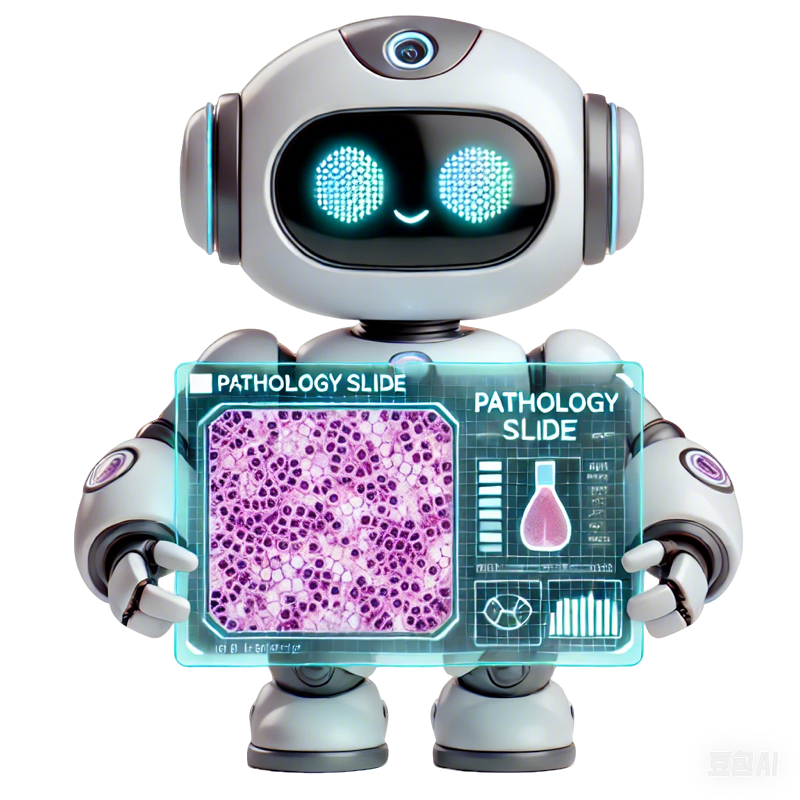} \ourmethod: A Collaborative Multi-Agent System for Multi-Modal Whole Slide Image Analysis}
\titlerunning{\ourmethod: A Multi-Agent System for Whole Slide Image Analysis}

%
\author{Xinheng Lyu\inst{1,2} % index{Lyu, Xinheng}
\and
Yuci Liang\inst{3} % index{Liang, Yuci}
\and
Wenting Chen\inst{4}$^\dagger$ % index{Chen, Wenting}
\and
Meidan Ding\inst{3} % index{Ding, Meidan}
\and
Jiaqi Yang\inst{2,3} % index{Yang, Jiaqi}
\and
Guolin Huang\inst{3,5} % index{Huang, Guolin}
\and
Daokun Zhang\inst{2} % index{Zhang, Daokun}
\and
Xiangjian He\inst{2}$^\dagger$ % index{He, Xiangjian}
\and
Linlin Shen\inst{1}$^\dagger$} % index{Shen, Linlin}
\institute{College of Artificial Intelligence, Shenzhen University, China \and
School of Computer Science, University of Nottingham Ningbo China, China \and
College of Computer Science and Software Engineering, Shenzhen University, China \and
Department of Electrical Engineering, City University of Hong Kong, Hong Kong \and
Wuyi University, China}
\renewcommand{\thefootnote}{\fnsymbol{footnote}}
\footnotetext{\scriptsize $^\dagger$Corresponding authors: Wenting Chen (wentichen7-c@my.cityu.edu.hk), Linlin Shen (llshen@szu.edu.cn), and Xiangjian He (Sean.He@nottingham.edu.cn)}
\maketitle

\begin{abstract}

Whole slide images (WSIs) are vital in digital pathology, enabling gigapixel tissue analysis across various pathological tasks. While recent advancements in multi-modal large language models (MLLMs) allow multi-task WSI analysis through natural language, they often underperform compared to task-specific models. Collaborative multi-agent systems have emerged as a promising solution to balance versatility and accuracy in healthcare, yet their potential remains underexplored in pathology-specific domains. To address these issues, we propose \ourmethod, a novel collaborative multi-agent system for multi-modal WSI analysis. \ourmethod~integrates specialized functional agents with robust task allocation and verification mechanisms to enhance both task-specific accuracy and multi-task versatility through three components: (1) a task allocation module assigning tasks to expert agents using a model zoo of patch and WSI level MLLMs, (2) a verification mechanism ensuring accuracy through internal consistency checks and external validation using pathology knowledge bases and domain-specific models, and (3) a summary module synthesizing the final summary with visual interpretation maps. Extensive experiments on multi-modal WSI benchmarks show \ourmethod's superiority to current WSI MLLMs and medical agent frameworks across diverse tasks. Source code is available at \url{https://github.com/XinhengLyu/WSI-Agents}.

\end{abstract}
\section{Introduction}

Whole slide images (WSIs) are essential in digital pathology, providing gigapixel-scale digitized tissue samples. WSI analysis has evolved from early patch-based models targeting specific tasks (cancer classification~\cite{hou2016patch,khened2021generalized}, grading~\cite{ccayir2023patch,bulten2020automated}, and tumor detection~\cite{khened2021generalized,ciga2021overcoming}) to advanced approaches handling gigapixel-scale analysis (survival prediction~\cite{tripathi2021end,hashimoto2020multi}, diagnosis~\cite{ding2024multimodal,xu2024whole}, report generation~\cite{guo2024histgen,chen2024wsicaption}, and visual question answering~\cite{chen2024wsi}). Despite impressive performance, most methods remain task-specific, limiting versatility. Consequently, there is a pressing need for comprehensive models capable of addressing multiple pathological tasks and adapting to diverse scenarios.

Recent advances in multi-modal large language models (MLLMs)\cite{liang2024wsi,chen2024slidechat} have enabled WSI-level MLLMs\cite{liang2024wsi,chen2024slidechat} capable of performing diverse pathological tasks through natural language interaction. Models such as WSI-LLaVA~\cite{liang2024wsi} and SlideChat~\cite{chen2024slidechat} handle various tasks including morphological analysis, diagnosis, treatment planning, and report generation. While offering multi-task capability within a single framework, they exhibit significant performance gaps compared to specialized foundation models. For example, while dedicated foundation models achieve over 90\% accuracy in cancer subtype classification, current MLLMs typically underperform by 15-30\% on comparable diagnostic tasks~\cite{liang2024wsi,chen2024slidechat,ding2024multimodal,shaikovski2024prism}. Therefore, improving task-specific performance while maintaining multi-task capabilities remains a critical challenge.

Recent AI agents~\cite{liu2023agentbench,liang2023encouraging,qian2023communicative,wang2024beyond} can leverage multiple specialized models and real-world APIs, offering a potential solution to the accuracy-versatility trade-off. In medical applications, some agents~\cite{kim2024mdagents,tang2023medagents,liu2024medchain} employ collaborative multi-agent systems for clinical reasoning, while others~\cite{fallahpour2025medrax,li2024mmedagent} integrate external machine learning tools to enhance diagnostic accuracy. However, these approaches often either specialize too narrowly (e.g., chest X-rays)~\cite{fallahpour2025medrax} or spread too broadly across imaging modalities~\cite{li2024mmedagent}, diluting domain expertise. Notably, these agents lack support for gigapixel WSIs, which are essential for comprehensive histopathological analysis. This limitation, coupled with insufficient pathology-specific knowledge, explains why general medical agents achieve only 40-50\% accuracy on pathology-specific tasks compared to 75-80\% on general medical tasks~\cite{kim2024mdagents,tang2023medagents}. Thus, incorporating both specialized pathology domain knowledge and robust WSI processing capabilities into medical agents is highly recommended for effective pathology analysis.

\begin{figure}[t]
    \centering
    \captionsetup{aboveskip=2pt, belowskip=2pt} 
    \includegraphics[width=0.82\textwidth]{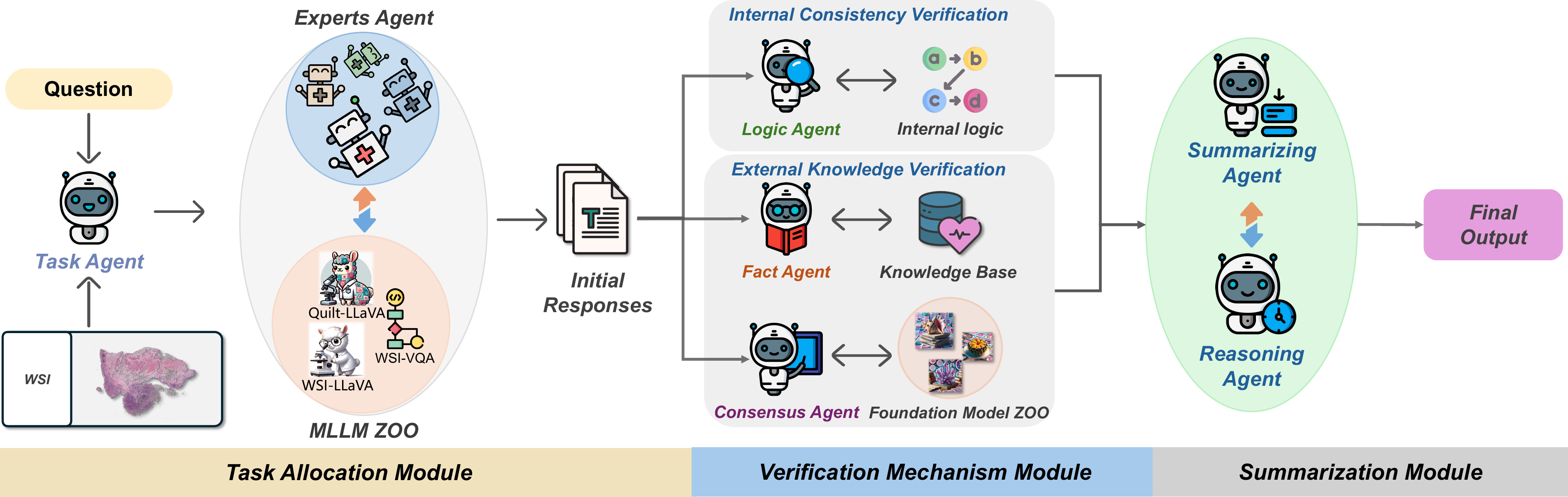} 
    \caption{The workflow of \ourmethod~with three main modules.}
    \label{fig:workflow}
    % \vspace{-0.4cm}
\end{figure}

To address these challenges, we propose \textbf{\ourmethod}, a novel collaborative multi-agent system for multi-modal WSI analysis that integrates specialized functional agents with robust verification mechanisms. As shown in Fig.\ref{fig:workflow}, \ourmethod consists of three primary components: a \textbf{task allocation module}, \textbf{verification mechanisms}, and a \textbf{summary module}. The task allocation module employs a \textbf{task agent} to interpret input tasks and assign them to specialized expert agents (morphology, diagnosis, report generation, etc.) who automatically select relevant models from our comprehensive model zoo, generating multiple initial responses.To ensure clinical accuracy, our \textbf{verification mechanism} includes \textbf{internal consistency verification} and \textbf{external knowledge verification}. Internal consistency verification uses a \textit{logic agent} to check for contradictions and evidence validity. More critically, external knowledge verification addresses the \textit{domain knowledge challenge} through fact and consensus agents. The \textit{fact agent} validates responses against pathology knowledge bases constructed from medical literature, while the \textbf{consensus agent} leverages domain knowledge from WSI foundation models pre-trained on histopathological datasets. Each verification agent generates scores and maintains detailed logs in the memory module. Finally, the \textit{summarizing agent} evaluates verification scores, selects optimal responses, and coordinates with \textit{reasoning agents} to produce the final output. For visual interpretation, we integrate attention maps from multiple patch-level WSI models into a comprehensive visual interpretation map. Extensive experiments on two major multi-modal WSI benchmarks demonstrate \ourmethod's superior performance compared to existing WSI MLLMs and medical agent frameworks.

\begin{figure}[t]
    \centering
    \includegraphics[width=1\textwidth]{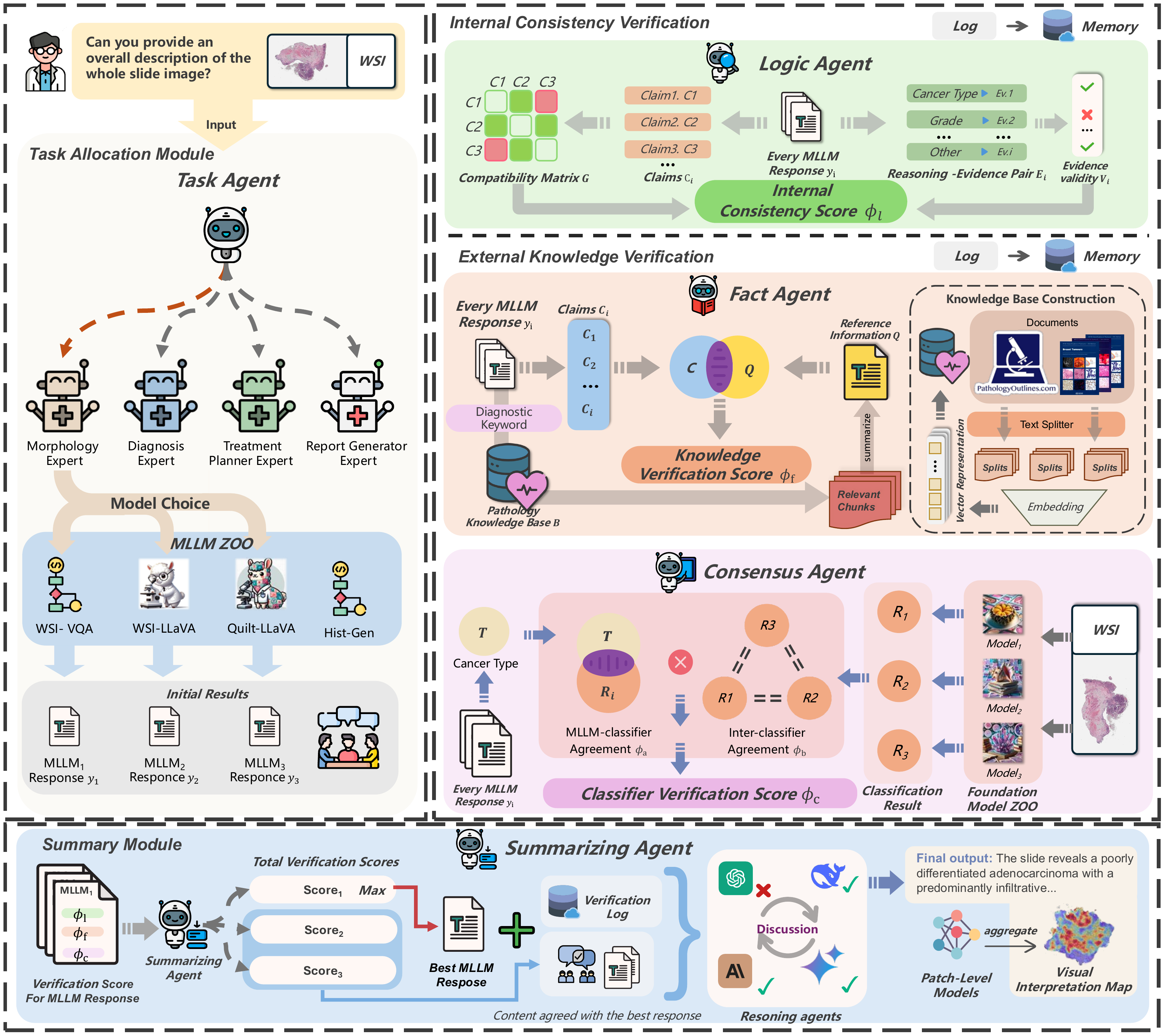} 
    \caption{Overview of \textbf{\ourmethod} for multi-modal WSI analysis with a \textbf{task allocation module} for expert allocation, \textbf{verification mechanisms} for accuracy assessment, and \textbf{summary module} for final summary.}
    \label{fig:lg_overview}
\end{figure}

\section{\ourmethod}
% \subsection{Overview}
\noindent\textbf{Overview.} The workflow of \ourmethod~is illustrated in Fig.~\ref{fig:lg_overview}, focusing on collaborative multi-agent WSI analysis. Given input WSI $I$ and question $x$, our system processes them through three modules: task allocation module, verification mechanism, and summary module. In the task allocation module, a task agent delegates responsibilities to specialized expert agents $A$ who select $M$ relevant models from our model zoo, generating multiple initial responses $\left\{y_i \right \}_{i=1}^M$. For verification, a logic agent extracts claims $\left\{C_i \right \}_{i=1}^N$ and evidences $\left\{E_i\right \}_{i=1}^N$, computes a compatibility matrix $G$ to represent the compatibility among claims, and evaluates evidence validity $\left\{V_i\right \}_{i=1}^N$ to determine internal consistency score $\phi_l$. The fact agent validates claims against a pathology knowledge base $B$ to compute verification score $\phi_f$. A consensus agent extracts diagnostic keywords $D$, compares them with foundation model results $\left\{R_i\right \}_{i=1}^H$ to compute MLLM-classifier agreement $\phi_a$, calculates inter-classifier agreement $\phi_b$ as a confidence factor, and multiplies these for classifier verification score $\phi_c$. In the summary module, a summarizing agent evaluates these verification scores and coordinates with reasoning agents to produce the final output $y$. For visualization, we integrate attention maps $\left\{m_i\right \}_{i=1}^K$ from $K$ patch-level models into a comprehensive interpretation map $m$, providing explainable visualization.

\subsection{Task Allocation Module (TAM)}

To adapt to diverse pathological scenarios, we introduce a task allocation module for WSI. This module comprises a task agent that interprets requirements and delegates specialized expert agents to generate preliminary responses, effectively addressing various pathology-specific challenges.

\noindent\textbf{Task Agent.} 
Specifically, given an input WSI $I$ and question $x$, the task agent first analyzes whether the question relates to WSI analysis, determines the task type of the given question, and identifies which specialized agents $A$ can correctly answer this question based on the task type. The task agent then assigns an appropriate specialized agent to address the question. 

\noindent\textbf{Expert Agent.} We design a series of specialized expert agents to address the primary tasks in WSI analysis, including a morphology expert for analyzing tissue architecture; a diagnosis expert for disease classification; a treatment planning expert for therapeutic recommendations and prognosis assessment; and a report generation expert for generating pathology reports. When assigned a task, the expert agent selects $M$ WSI MLLMs from a pre-defined MLLM model zoo, choosing models that have been trained on relevant tasks and can address the question, and leverages them to generate initial responses $\left\{y_i \right \}_{i=1}^M$. We construct the model zoo by integrating five WSI analysis MLLMs~\cite{chen2024wsi,seyfioglu2024quilt,liang2024wsi,chen2024wsicaption,guo2024histgen}.

\subsection{Internal Consistency Verification (ICV)}
To ensure the logical consistency of generated responses, we introduce an internal consistency verification system that leverages a logic agent to check for contradictions and validate evidence within responses.

\noindent\textbf{Logic Agent.} Concretely, the logic agent first extracts the claims $\{C_i\}_{i=1}^N$ from the initial responses $y_i$, where $N$ denotes the number of claims. To analyze contradictions within the response, we compute a compatibility matrix $G \in \mathbb{R}^{N \times N}$ representing the compatibility among different claims. Each element $G_{ij} \in [0,1]$ indicates whether claim $C_i$ can coexist with claim $C_j$. We then compute the compatibility score $\phi_g$ to represent the consistency among claims:
\begin{equation}
\phi_g = \frac{\sum_{i=1}^{N}\sum_{j=i+1}^{N} G_{ij}}{N(N-1)/2}.
\end{equation}
To further validate each individual claim, the logic agent extracts evidence $\{E_i\}_{i=1}^N$ from the response for each claim and assesses the evidence validity $\{V_i\}_{i=1}^N \in [0,1]$ to verify whether evidence $E_i$ supports its corresponding claim $C_i$. The validity score $\phi_e$ is calculated as $\phi_e = \frac{\sum_{i=1}^N V_i}{N}$. The final internal consistency score $\phi_l$ is the average of these two scores, 
$\phi_l = \frac{\phi_g + \phi_e}{2}$.

\subsection{External Knowledge Verification (EKV)}
Due to the limited pathological knowledge in current medical agents \cite{kim2024mdagents,tang2023medagents,liu2024medchain,fallahpour2025medrax,li2024mmedagent}, we design an external knowledge verification system that employs a fact agent to check for factual errors using a comprehensive knowledge base, alongside a consensus agent that identifies agreement with knowledge embedded in WSI foundation models.

\noindent\textbf{Fact Agent.} To check for factual errors in generated responses, the fact agent first extracts claims $\{C_i\}_{i=1}^N$ from the responses and selects their diagnostic keywords $k$. To obtain relevant knowledge about the input question and response, we establish a comprehensive pathology knowledge base $B$ by collecting resources from pathology websites~\cite{pathWeb} and authoritative literature~\cite{nagtegaal2019,tan20202019,bosman2010classification}. To facilitate efficient retrieval, we segment the documents into splits, extract embeddings for each split, and implement indexing for similarity-based retrieval. Using $k$, we retrieve relevant chunks from $B$ and summarize them into reference information $Q$. To assess factual consistency, we compute a factual score $f_i$ for each claim by determining whether $C_i$ violates the knowledge in $Q$, then average these scores to obtain the knowledge verification score $\phi_k = \frac{\sum_{i=1}^N f_i}{N}$. 

\noindent\textbf{Consensus Agent.}
Since recent WSI foundation models~\cite{ding2024multimodal,lu2024visual,lu2023visual} have acquired extensive knowledge through self-supervised learning on large-scale datasets, we design a consensus agent to leverage this embedded expertise and identify agreement between these models and generated responses. The consensus agent first extracts the cancer type $T$ from the generated responses. It then draws upon a foundation model zoo comprising several state-of-the-art WSI foundation models (TITAN~\cite{ding2024multimodal}, CONCH~\cite{lu2024visual}, and Prism~\cite{shaikovski2024prism}), each demonstrating exceptional classification performance. Next, the agent obtains classification results $\{R_i\}_{i=1}^H$ from these models, where $H$ represents the number of classifiers. The consensus agent calculates the agreement $A_i$ between the extracted cancer type $T$ and each classification result $\{R_i\}_{i=1}^H$, then computes the MLLM-classifier agreement $\phi_a = \frac{\sum_{i=1}^H A_i}{H}$. To account for reliability among different classifiers, the agent also determines the inter-classifier agreement $\phi_b$ using:
\begin{equation}
\phi_b = \frac{2\sum_{i=1}^{H}\sum_{j=i+1}^{H}\mathbbm{1}(R_i=R_j)}{H(H-1)},
\end{equation}
where $\mathbbm{1}(\cdot)$ is an indicator function. The final classifier verification score $\phi_c$ is calculated by using the inter-classifier agreement as a confidence factor: $\phi_c = \phi_a \cdot \phi_b$. 
The verification logs of logic, fact and consensus agents are stored in the memory module.

\subsection{Summary Module}
\noindent\textbf{Summarizing Agent.} A summarizing agent synthesizes the final results by unifying all verification scores into a total verification score, $\phi_{total} = w_1\phi_l + w_2\phi_k + w_3\phi_c$, where $w_1$, $w_2$, and $w_3$ represent the importance weights of different verification processes. The agent then identifies the MLLM response with the highest score as the best response and extracts content from other responses that align with this selection. Using this best response, along with the extracted content and verification logs, the summarizing agent generates an initial summary.

\noindent\textbf{Reasoning Agents.} To further enhance the process, reasoning agents evaluate the initial summary and provide their expert opinions. When disagreements arise, these agents submit specific revision suggestions. The summarizing agent then incorporates this feedback to generate a revised summary for the next round of discussion. This iterative deliberation continues until a consensus is reached—specifically, when more than half of reasoning agents endorse the summary—at which point the discussion concludes and produces final summary. For visual interpretation, we integrate attention maps from multiple patch-level WSI models and synthesize them into a comprehensive visual interpretation map.% $m$.

\begin{table}[h!]
\centering
\footnotesize
\caption{Quantitative comparison on WSI-Bench dataset.}
\setlength{\tabcolsep}{2pt} 
\scalebox{0.82}{ 
\begin{tabular}{cccccccccccccccc}
\toprule
\multirow{3}{*}{Methods} & \multicolumn{5}{c}{\textbf{Morph. Analysis}} & \multicolumn{5}{c}{\textbf{Diagnosis}} & \multicolumn{3}{c}{\textbf{Treat. Plan.}} & \multirow{3}{*}{Avg.} \\ \cmidrule(lr){2-6} \cmidrule(lr){7-11} \cmidrule(lr){12-14}
 & \makecell{G.M.} & \makecell{K.D.} & \makecell{R.S.} & \makecell{S.F} & \makecell{Avg.} & \makecell{H.T.} & \makecell{G.R.} & \makecell{M.S.} & \makecell{S.T.} & \makecell{Avg.} & \makecell{T.R.} & \makecell{P.R.} & \makecell{Avg.} & \\
\midrule
Quilt-LLaVA\cite{seyfioglu2024quilt} & 0.338 & 0.314 & 0.389 & 0.752 & 0.448 & 0.339 & 0.505 & 0.675 & 0.824 & 0.586 & 0.764 & 0.812 & 0.788 & 0.607 \\
WSI-VQA\cite{chen2024wsi} & 0.322 & 0.313 & 0.389 & 0.554 & 0.395 & 0.377 & 0.430 & 0.388 & 0.550 & 0.436 & 0.708 & 0.874 & 0.791
 & 0.541 \\
WSI-LLaVA\cite{liang2024wsi} & 0.390 & 0.350 & 0.450 & 0.760 & 0.488 & 0.410 & 0.570 & 0.630 & 0.830 & 0.610 & 0.790 & 0.830 & 0.810 & 0.610 \\
\midrule
MDAgents\cite{kim2024mdagents} & 0.075 & 0.147  &0.224 & 0.460  & 0.227 & 0.038 & 0.315  &  0.487 & 0.125 & 0.241 & 0.611 & 0.405 & 0.508 & 0.326  \\
Med-Agents\cite{tang2023medagents} & 0.199 & 0.288 & 0.319 & 0.615 &0.356 & 0.169 & 0.311 & 0.419 & 0.700 & 0.400 & 0.775 & \textcolor{blue}{0.876} & 0.825 & 0.528  \\
\midrule
\textbf{WSI-Agents} & \textcolor{blue}{0.512} & \textcolor{blue}{0.440} & \textcolor{blue}{0.534} & \textcolor{blue}{0.786} & \textcolor{blue}{0.568} & \textcolor{blue}{0.529} & \textcolor{blue}{0.586} & \textcolor{blue}{0.840} & \textcolor{blue}{0.900} & \textcolor{blue}{0.714} & \textcolor{blue}{0.823} & 0.831 & \textcolor{blue}{0.827} & \textcolor{blue}{0.703} \\
\bottomrule
 \end{tabular}}
 \tiny Abbreviation: G.M. (Global Morphological Description), K.D. (Key Diagnostic Description), R.S. (Regional Structure Description), S.F. (Specific Feature Description), H.T. (Histological Typing), 
G.R. (Grading), M.S. (Molecular Subtyping), S.T. (Staging), T.R. (Treatment Recommendations), P.R. (Prognosis).
\label{tab:wsi_bench}
% \vspace{-0.1cm}
\end{table}

\begin{table}[ht!]
    \centering

    % Begin first minipage for the first table
    \begin{minipage}{0.58\textwidth}
        \centering
        % \footnotesize
        % \scriptsize
        % \setlength\tabcolsep{1.1pt}%{0.5pt}
        \caption{\small Comparison on report generation task.}
    \label{tab:report_gen}
        \scalebox{0.8}{
	    \begin{tabular}{lccccccc}
	        \toprule
	        Models & B-1 & B-2 & B-3 & B-4 & R-L & M & Acc \\ 
	        \midrule
	        Quilt-LLaVA\cite{seyfioglu2024quilt} & 0.3343 & 0.1698 & 0.0928 & 0.0566 & 0.2463 & 0.2910 & 0.324 \\ 
	        MI-Gen\cite{chen2024wsicaption} & 0.4027 & 0.3061 & 0.2481 &\textcolor{blue}{0.2085} & 0.4464 & 0.4070 & 0.310 \\ 
	        Hist-Gen\cite{guo2024histgen} & 0.4058 & 0.3070 & \textcolor{blue}{0.2482} & 0.2081 &\textcolor{blue}{0.4484} & 0.4162 & 0.300 \\ 
	        WSI-LLaVA\cite{liang2024wsi} & 0.3531 & 0.1859 & 0.1058 & 0.0665 & 0.2626 & 0.3072 & 0.380 \\ \hline
	        MDAgents\cite{kim2024mdagents}& 0.1287 & 0.0521 & 0.0219 & 0.0118 & 0.1572 &  0.1623 & 0.105  \\ 
	        Med-Agents\cite{tang2023medagents}&0.1610  &0.0659  &0.0255  &  0.0130& 0.1997 &0.2273  & 0.174 \\ \hline
	        \textbf{\ourmethod} & \textcolor{blue}{0.4443} & \textcolor{blue}{0.3084} & 0.2307 & 0.1825 & 0.4479 & \textcolor{blue}{0.4719} & \textcolor{blue}{0.440} \\ 
	        \bottomrule
	    \end{tabular}
        }
        \scriptsize 
        \textit{Abbreviations:} B-1/B-2/B-3/B-4 (BLEU-1/2/3/4), R-L (ROUGE-L), M (METEOR), Acc (WSI-Precision~\cite{liang2024wsi}).
    \end{minipage}
    \hfill
    \begin{minipage}{0.32\textwidth}
        \centering
        % \footnotesize 
        % \scriptsize
        \caption{\small Quantitative Evaluation on WSI-VQA.}
    \label{tab:wsi_vqa}
        \scalebox{0.8}{ 
		    \begin{tabular}{lc}%ccccc}
		        \toprule
		        Models & Acc \\ \hline
                Quilt-LLaVA\cite{seyfioglu2024quilt} & 0.130 \\
                WSI-VQA\cite{chen2024wsi} & 0.470 \\
                WSI-LLaVA\cite{liang2024wsi}  & 0.550 \\ \hline
                MDAgents\cite{kim2024mdagents} & 0.208 \\
                Med-Agents\cite{tang2023medagents} & 0.183 \\ \hline
                \textbf{\ourmethod} & \textcolor{blue}{0.600} \\ 
		        \bottomrule
    \end{tabular}
    }
    \scriptsize 
    \textit{Abbreviations:} Acc (Accuracy).
    \end{minipage}
\end{table}
\section{Experiments}

\noindent\textbf{Datasets and Implementation.} We conduct experiments on two main WSI benchmarks, i.e., WSI-Bench~\cite{liang2024wsi} and WSI-VQA~\cite{chen2024wsi} datasets, and follow their evaluation protocol. We use AutoGen~\cite{autogen} to build our agent framework. For the model with limited input size, all the WSIs are resized to $1024\times1024$ thumbnails.For better evaluation of open-ended questions, we employ WSI-Precision, which compares extracted claims from ground-truth with model outputs and averages the scores; this metric is used in Table~\ref{tab:wsi_bench} and as the accuracy (Acc) measure in Table~\ref{tab:report_gen}.

\noindent\textbf{Results on WSI-Bench.}
We evaluate \ourmethod~against WSI MLLMs~\cite{seyfioglu2024quilt,chen2024wsi,liang2024wsi} and medical agents~\cite{kim2024mdagents,tang2023medagents} on the WSI-Bench dataset~\cite{liang2024wsi}. Table~\ref{tab:wsi_bench} demonstrates that \ourmethod~achieves superior performance across all three main tasks, surpassing Quilt-LLaVA~\cite{seyfioglu2024quilt} and Med-Agents~\cite{tang2023medagents} by substantial margins of approximately 10\% and 17\%, respectively. For the report generation task, Table~\ref{tab:report_gen} shows that \ourmethod~outperforms WSI-LLAVA~\cite{liang2024wsi} by 6\% in accuracy. These results clearly indicate that \ourmethod~surpasses existing WSI MLLMs and medical agents in performance. For qualitative comparison, Table~\ref{tab:qual_vqa} shows other methods misclassify the sample despite identifying some correct features, while \ourmethod~correctly identifies adenocarcinoma with accurate supporting evidence, demonstrating superior diagnostic precision in WSI interpretation.

\noindent\textbf{Results on WSI-VQA Dataset.}
In Table~\ref{tab:wsi_vqa}, our method achieves 60.0\% accuracy, outperforming previous approaches including WSI-LLaVA~\cite{liang2024wsi} (55.0\%), WSI-VQA~\cite{chen2024wsi} (47.0\%), and Quilt-LLaVA~\cite{seyfioglu2024quilt} (13.0\%). Medical agents like MDAgents~\cite{kim2024mdagents} (20.8\%) and Med-Agents~\cite{li2024mmedagent} (18.3\%) performed notably worse, suggesting general medical knowledge is insufficient for WSI analysis. These results highlight the superiority of \ourmethod~through its effective integration of pathology knowledge and verification mechanism.

\begin{table}[h!]

\caption{Visual comparison on WSI-Bench. \textcolor{green!50!black}{Green} for correct and \textcolor{red!50!black}{Red} for wrong.}
\scriptsize
\begin{tabular}{m{1.3cm} m{11cm}} 
% \scalebox{0.75}{
\toprule[1.5pt]
\multirow{2}{*}{\textbf{Image}} &
\begin{minipage}{11cm}
    \centering
    \begin{tabular}{cc}
        \includegraphics[width=2.3cm,height=2cm]{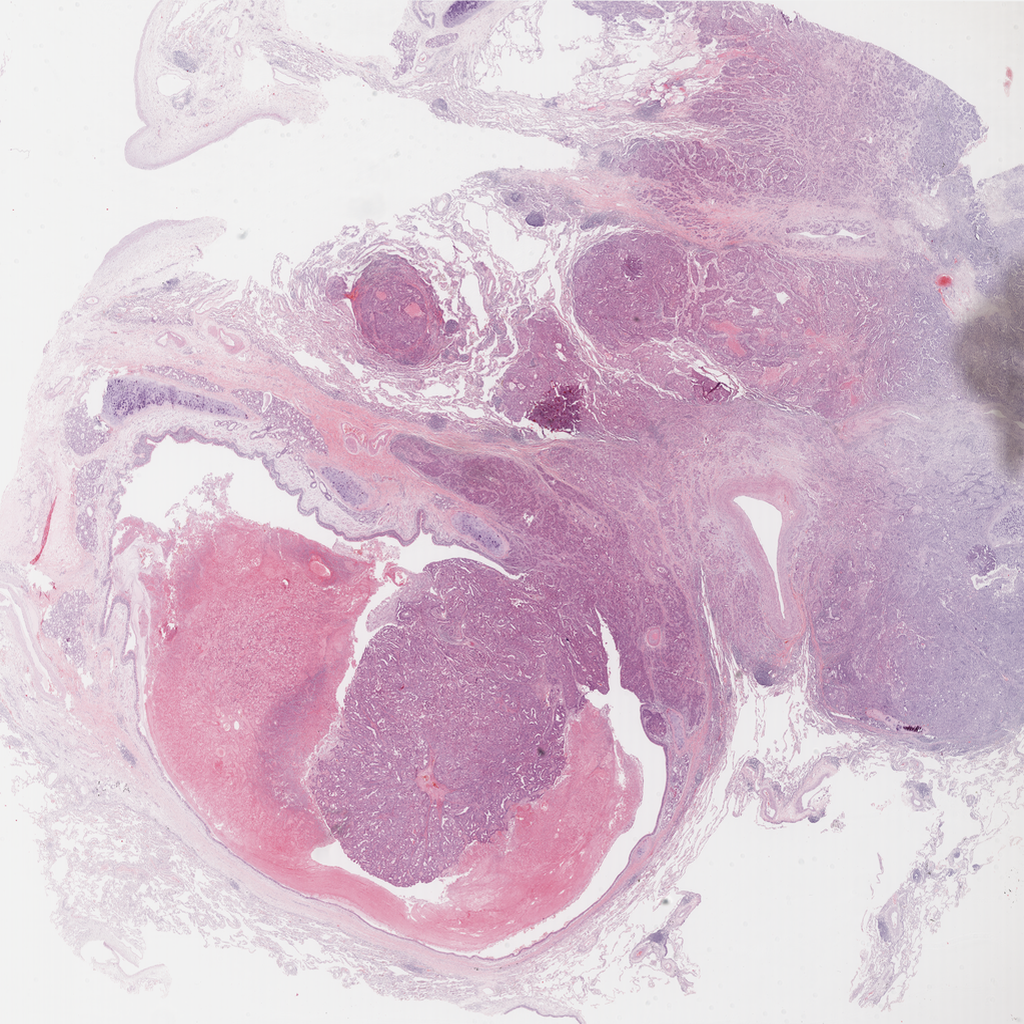} & 
        \includegraphics[width=2.3cm,height=2cm]{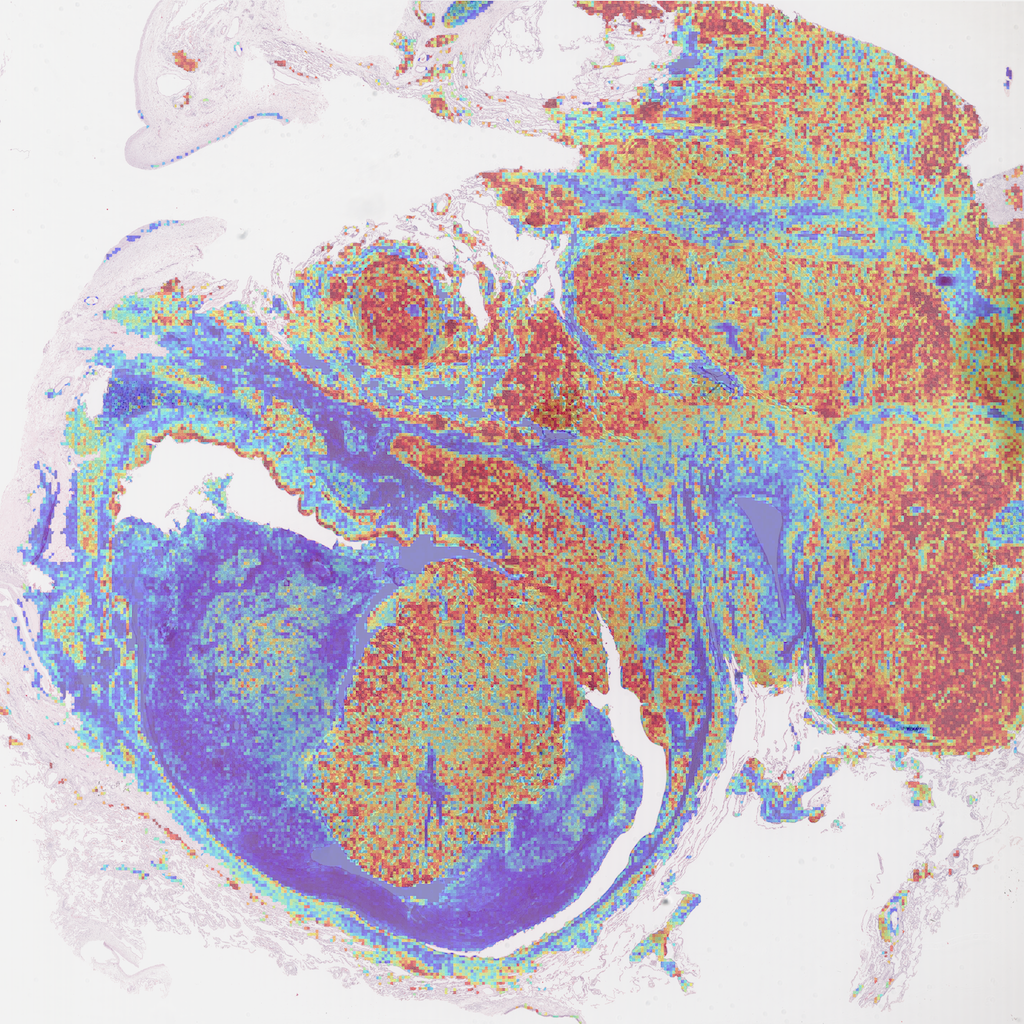} \\
        \scriptsize Input WSI & \scriptsize Generated Visual Interpretation Map
    \end{tabular}
\end{minipage} \\
\midrule
\textbf{Question} & What is the histological classification based on your examination of the slide? \\

\midrule
\textbf{Quilt-LLaVA} & The histological classification is \textcolor{red!50!black}{seminoma}. This is supported by the presence of \textcolor{red!50!black}{large, uniform cells with clear cytoplasm and prominent nucleoli, arranged in nests and sheets, and the characteristic fibrous stroma with lymphocytic infiltration}. \\
\midrule
\textbf{WSI-VQA} & The histological classification based on the observed features is \textcolor{red!50!black}{invasive squamous cell carcinoma} . This classification is supported by the presence of \textcolor{green!50!black}{poorly differentiated }\textcolor{red!50!black}{squamous cells} with \textcolor{green!50!black}{marked nuclear pleomorphism and high mitotic activity}.
 \\
\midrule
\textbf{WSI-LLaVA} & The histological classification is \textcolor{red!50!black}{high-grade urothelial carcinoma}. This classification is supported by the presence of \textcolor{green!50!black}{poorly differentiated tumor cells, significant nuclear atypia, high mitotic activity, and evidence of lymphovascular invasion}. \\
\midrule
\textbf{\ourmethod} & The histological classification...is \textcolor{green!50!black}{adenocarcinoma}. This classification is consistent with the identification of \textcolor{green!50!black}{glandular structures,...poorly differentiated cells with nuclear pleomorphism and high mitotic activity}. Additionally,...keratinization or intercellular bridges...are absent in the provided image.There is no evidence of lymphovascular invasion or specific stromal characteristics...were noted in the image, their absence aligns with the adenocarcinoma classification rather than alternatives like squamous cell carcinoma, high-grade urothelial carcinoma or seminoma. \\
\midrule
\textbf{Ground Truth} & Based on the observed features, the histological classification is adenocarcinoma. The poorly differentiated nature of the glandular epithelial cells, along with marked pleomorphism and significant mitotic activity, supports this classification. The invasive pattern further confirms the diagnosis of adenocarcinoma. \\ 
\bottomrule[1.5pt]

\end{tabular}
\label{tab:qual_vqa}
\end{table}

\begin{table}[ht!]
    \centering
    \caption{Ablation studies on WSI-Bench dataset.}
    \footnotesize
        \scalebox{0.8}{
        \begin{tabular}{ccccccccccccc}
            \toprule

            \multicolumn{2}{c}{TAM} & ICV& \multicolumn{2}{c}{EKV} & \multicolumn{2}{c}{Summary}  & \multirow{2}{*}{\makecell{Morph.\\Analysis}}&\multirow{2}{*}{Diagnosis} &\multirow{2}{*}{\makecell{Treat.\\Plan.}} &\multirow{2}{*}{\makecell{Report\\Gen.}} &\multirow{2}{*}{Avg.} \\ \cmidrule(lr){1-2} \cmidrule(lr){3-3} \cmidrule(lr){4-5} \cmidrule(lr){6-7} 
            Task & 
            Expert  & 
            Logic  & 
            Fact  & 
            Cons.  & 
            Summ.  & 
            Reas.   & 
             & 
             & 
             & 
             & 
             \\
            \midrule
            \ding{55}& \ding{51}& \ding{51} & \ding{51} & \ding{51} & \ding{51}& \ding{51}&  0.362&0.560  & 0.786&0.415  & 0.531 \\ 
            \ding{51}& \ding{55}& \ding{51} & \ding{51} & \ding{51} & \ding{51}& \ding{51}&0.289  &0.427&  0.528& 0.321  &0.391  \\ 
           \ding{51}&\ding{51} & \ding{55} & \ding{51} & \ding{51} & \ding{51}&\ding{51} & 0.539 & 0.648 & 0.817 & 0.424 & 0.607 \\ 
           \ding{51}& \ding{51}&  \ding{51} & \ding{55} & \ding{51}  &\ding{51} & \ding{51}& 0.521 & 0.647 & 0.814 & 0.431 & 0.603 \\ 
            \ding{51}& \ding{51}& \ding{51} & \ding{51} & \ding{55} & \ding{51}& \ding{51}& 0.531 & 0.644 & 0.814 & 0.418 & 0.601 \\ 
            \ding{51}& \ding{51}& \ding{51} & \ding{51} & \ding{51} & \ding{55}& \ding{51}&0.416&  0.586&  0.810& 0.332&0.536    \\ 
            \ding{51}& \ding{51}& \ding{51} & \ding{51} & \ding{51} & \ding{51}& \ding{55}&0.469  &0.607&  0.806& 0.362&  0.561  \\ 
            \ding{51}& \ding{51}& \ding{51} & \ding{51} & \ding{51} & \ding{51}& \ding{51}& 0.568 & 0.714 & 0.827 & 0.440 & 0.637 \\ 
            \bottomrule
        \end{tabular}}

    \label{tab:ablation_studies}
    % \vspace{-0.1cm}
 \tiny {Abbreviations:} TAM (Task Allocation Module), ICV (Internal Consistency Verification), EKV (External Knowledge Verification), Cons. (Consensus), Summ. (Summarizing), Reas. (Reasoning), Morph. (Morphological), Treat. Plan. (Treatment Planning), Report Gen. (Report Generation), Avg. (Average).
\end{table}

\noindent\textbf{Ablation Studies.}
In Table~\ref{tab:ablation_studies}, we conduct ablation studies on the WSI-Bench dataset to evaluate the effectiveness of each module, including the task allocation module (TAM), internal consistency verification (ICV), external knowledge verification (EKV) and summary module (summary). When ablating each agent in the corresponding module, the performance across all the tasks decreases significantly, indicating the effectiveness of each agent.

\section{Conclusion}
We present a collaborative multi-agent framework (\ourmethod) for multi-modal WSI analysis by integrating specialized agents and verification mechanisms. Through task allocation, consistency checks, knowledge verification, and summarization, \ourmethod~improves accuracy, adaptability, and interpretability for pathological tasks. Experiments demonstrate superior performance over existing WSI MLLMs and current medical agents. Ablation studies prove the effectiveness of each module. 

\begin{credits}
\subsubsection{\ackname} This study was supported by the National Natural Science Foundation of China (Grants 82261138629 and 12326610), the Guangdong Provincial Key Laboratory (Grant 2023B1212060076), the Yongjiang Technology Innovation Project (Grant 2022A-097-G), the Zhejiang Department of Transportation General Research and Development Project (Grant 2024039), and the National Natural Science Foundation of China talent grant (UNNC: B0166).

\subsubsection{\discintname}
The authors have no competing interests to declare that are relevant to the content of this article.
\end{credits}


\end{document}